\documentclass{article}

\usepackage{microtype}
\usepackage{graphicx}
\usepackage{subfigure}
\usepackage{booktabs} % for professional tables

\usepackage{hyperref}

\usepackage[accepted]{icml2023_dp4ml}

\usepackage{amsmath}
\usepackage{amssymb}
\usepackage{mathtools}
\usepackage{amsthm}
\usepackage{natbib}
\usepackage{svg}
\usepackage{caption}

\usepackage[capitalize,noabbrev]{cleveref}

\theoremstyle{plain}

\theoremstyle{definition}

\theoremstyle{remark}

\usepackage[textsize=tiny]{todonotes}

\icmltitlerunning{A convex-concave spline approximation of neural networks}

\begin{document}

\twocolumn[
\icmltitle{A max-affine spline approximation of neural networks using the Legendre transform of a convex-concave representation}

% It is OKAY to include author information, even for blind
% submissions: the style file will automatically remove it for you
% unless you've provided the [accepted] option to the icml2023
% package.

% List of affiliations: The first argument should be a (short)
% identifier you will use later to specify author affiliations
% Academic affiliations should list Department, University, City, Region, Country
% Industry affiliations should list Company, City, Region, Country

% You can specify symbols, otherwise they are numbered in order.
% Ideally, you should not use this facility. Affiliations will be numbered
% in order of appearance and this is the preferred way.
\icmlsetsymbol{equal}{*}

\begin{icmlauthorlist}
\icmlauthor{Adam Perrett}{yyy}
\icmlauthor{Danny Wood}{yyy}
\icmlauthor{Gavin Brown}{yyy}
%\icmlauthor{}{sch}
%\icmlauthor{}{sch}
\end{icmlauthorlist}

\icmlaffiliation{yyy}{Department of Computer Science, The University of Manchester, Manchester, UK}
% \icmlaffiliation{comp}{Company Name, Location, Country}
% \icmlaffiliation{sch}{School of ZZZ, Institute of WWW, Location, Country}

\icmlcorrespondingauthor{Adam Perrett}{adam.perrett@manchester.ac.uk, adamperrett3142@gmail.com}
% You may provide any keywords that you
% find helpful for describing your paper; these are used to populate
% the "keywords" metadata in the PDF but will not be shown in the document
\icmlkeywords{Machine Learning, ICML}

\vskip 0.3in
]

\printAffiliationsAndNotice{}  % leave blank if no need to mention equal contribution
% \printAffiliationsAndNotice{\icmlEqualContribution} % otherwise use the standard text.

\begin{abstract}
This work presents a novel algorithm for transforming a neural network into a spline representation. Unlike previous work that required convex and piecewise-affine network operators to create a max-affine spline alternate form, this work relaxes this constraint. The only constraint is that the function be bounded and possess a well-define second derivative, although this was shown experimentally to not be strictly necessary. It can also be performed over the whole network rather than on each layer independently. As in previous work, this bridges the gap between neural networks and approximation theory but also enables the visualisation of network feature maps. Mathematical proof and experimental investigation of the technique is performed with approximation error and feature maps being extracted from a range of architectures, including convolutional neural networks.

% %iteration
% This work applies the Legendre transform to a neural network and samples using the training data to create a max-affine spline (MAS) representation of a neural network enabling highly parallel inference. This opens doors to approximation theory and understanding network complexity. Unlike previous work that required activation functions to be piecewise-affine or
% convex operators and must applied to each layer individually this work requires only that the second derivative be well-defined, although, it will be shown this is not necessarily a requirement in creating a MAS representation. 
% %method
% By calculating the minimum and maximum eigenvalues of the Hessian of the neural network around the training data a convex and concave component can be added to the network which splits the network into a convex and concave pair. This allows separate MAS to be created of each convex and concave element by sampling from the Legendre transform around the training data points, creating a Convex-Concave-Spline (CCS) representation. 
% %results
% Following subsampling of planes that make up the MAS a reduced computational version can be extracted with minimal degradation of the functional quality. Various techniques for subsampling are contrasted. Key features and network functionality are explored with the MAS representation as well as comparison with piece-wise linear approximation. 

\end{abstract}

\section{Introduction}
% network interpretability
% Model understanding, explanation and interpretation
% feature extraction
% links to approximation theory

% network size
% over-parameterised
% computational understanding
% inference cost
\subsection{Interpreting networks}
% interpreability, medical cars, black box->light box, validation
Interpretability of neural networks is of increasing importance as they become embedded in more facets of everyday life. Applications such as medical diagnoses or self-driving cars would greatly benefit from easily applied inspection techniques. This would enable extracted network features to be validated and possibly used to make inferences about the data, such as knowing what features are indicative of certain diseases and responses. Being able to shine light into the black boxes that are neural networks facilitates verification beyond validation with hidden test data.

% gd for features, lots of inferences, issues with the inputs
A common technique for understanding what input features correspond to particular outputs is to apply gradient ascent on the inputs whilst aiming to maximise a particular output's probability, also termed input optimisation~\cite{Erhan2009visGA, Simonyan2013visGAvsdeconv}. As it relies on gradient techniques it can fall into common issues such as gradient saturation, initialisation sensitivity and local optima. There is also little guarantee about the extracted saliency map's relation to the trained features of the network. An alternative approach is the use of a complimentary deconvolutional network~\cite{Zeiler2014visualizing, Linardatos2020InterpretabilityReview}. In effect the same filters of a convolutional layer are applied in reverse to extract the input saliency map. They have shown good performance in network interpretability but lack spatial precision because of the reverse convolution operation. They are also architecture dependant and therefore not a general purpose tool. Techniques are also available to understand what features were selected as pertinent in an image using gradient information~\cite{Zhou2016saliency, Selvaraju2017gradcam}, although this is on an image-by-image basis and does not extract general properties of the trained network.

% \subsection{More parameters than needed}
% % over-parameterised, regularisation, ineffcient compute
% With the increased efficacy of regularisation techniques~\cite{?} and availability of computation resource it is common practice to over-parameterise networks~\cite{?}. This removes any bottleneck in performance caused by limited parameters without sacrificing performance thanks to regularisation techniques, such as dropout~\cite{?}. This creates inefficient utilisation of computation resource as many more parameters are processed than is strictly necessary~\cite{?}.

% % network distillation, training from scratch
% Network distillation~\cite{?} offers an approach to extracting the learnt knowledge of a large model and condensing it into a smaller model. The large model leverages its abundant parameters to learn the probability distribution over the data. The large model's output is then used as a soft label for training data when teaching the small model which can benefit from the increased information present, e.g. knowing the input looks like a 3 and an 8 rather than just knowing the input is a 3. This can significantly reduce the number of parameters with little reduction in performance. The main downside is the need for training another network from scratch and the correct application of training schedule, such as how soft the labels are at what point of the training.

\subsection{Network transformation}
% mention madmax, explain MAS
It was previously shown by Balestriero et al. that neural networks could be represented as a combination of Max-Affine Spline Operators (MASO)~\cite{Balestriero2018spline, Balestriero2021MASO}. Splines are a way of approximating a continuous function with a combination of straight lines, or planes in high dimensions. This work provided a link between deep neural networks and approximation theory and was used to generate a regularisation technique which aided training. 
% The equations of the planes and also `knots', which determine how planes interact, are required create a continuous function. Max-affine splines are a special subset of splines which model convex functions. The requirement to set knots is removed by taking the maximum value of all planes, thus why it only works for convex functions as the gradient must always increase as shown in Figure~\ref{?}. 
% explain madmax and its limitations
The contribution of Balestriero et al. was to show that neural network layers using convex operators, including convolution and pooling, can be represented as max-affine splines if they are piecewise-affine, such as ReLU, and approximated arbitrarily close if they are not piecewise. There is also discussion of how non-convex operators could be modelled as a combination of MASOs, although this is not explored experimentally. This technique is limited to creating a MASO of each layer individually and cannot be used to extract feature maps of the whole network.

The work of this paper presents a novel method for network transformation that converts all layers simultaneously into a pair of complementary Convex and Concave Splines (CCS). This alternate form allows feature maps to be extracted from the trained network to aid interpretability. There are no requirements of convexity, it only needs to be a bounded function with a well-defined second derivative, although it will be show experimentally how this constraint can be relaxed. 

\subsection{Contributions}

\begin{enumerate}
    \item A novel method of splitting a neural network into complementary convex and concave components
    \item The application and subsequent sampling of a Legendre transform of the convex and concave components to create a max-affine spline approximation of the original neural network that is embarrassingly parallel in operation
    \item A novel method for feature visualisation and interpretation is proposed
    \item The investigation of the CCS approximate form to evaluate network complexity
\end{enumerate}

% \subsection{The bigger picture}
% Networks are growing in size and as a results are becoming uninterpretable
% The over-parameterisation of neural networks leads to computational inefficiency

% Techniques need to be developed to further understanding of complex networks and reduce inference cost

% \subsection{Motivation}
% Techniques to inspect require gradient descent (similar to adversarial stuff)
% Or have certain limitations on network structure (like madmax)
% Network distillation requires retraining of another network

% There are many important techniques from approximation theory
% Transforming a network can allow different analysis

% \subsection{Where the work fits in}
% By constructing a convex and concave representation which combined recreates the original function:
% Legendre transform and sampling via data points creates MAS
% Samples can be taken which approximate the original function
% They can reduced down to constituent parts
% A max-affine-spline representation links NN to approximation theory

% \subsection{The results}
% An x\% reduced computational load with x reduction in accuracy
% All networks are a combination of a convex and concave function
% Potentially makes optimisation easier?
% Works for complex/non-convex/non-piecewise activation functions
% Different sampling methods have been explored

\section{Background}
% methods for inspection - probing
% relation to splines
    % ReLU
    % theoretical understanding
% splines for network inspection
% a need for knots
% concave-convex problems
Inspection of neural network behaviour is of increasing importance as they become crucial components of safety critical functions. There have been probing techniques which inspect the network~\cite{Zeiler2014visualizing, Linardatos2020InterpretabilityReview}. Common techniques involve performing gradient ascent on the inputs to create representation that maximises the activation of hidden neurons or outputs. There have also been investigations into the theoretical bounds of neural network capacity, although these fall short of explaining a trained neural network's behaviour~\cite{Tishby2015bottleneck, Achille2018representations}. An alternative neural network spline representation was presented by Balestriero et al.~\cite{Balestriero2018spline, Balestriero2021MASO} which showed that, by converting a network, properties can be inspected and understood using established mathematical techniques from approximation theory. Other authors have made the link between neural networks and splines but often only for single layers and in combination with piecewise linear (PWL) activations, such as ReLU~\cite{Parhi2022splines, DeVore2021NNapprox, Bunel2017PiecewiseNN, Bunel2020branchPWL, Misener2010PWLapprox, Chu2018PWLNN}. Theoretical extensions are suggested for non-piecewise activation functions but to the authors knowledge there are no experimental examples as of writing.

\subsection{Affine (Linear) splines}
Splines make an attractive alternative form for neural networks with PWL activations as it becomes possible to exactly match the original function with a finite set of splines and they also map $\mathbb{R}^D \rightarrow \mathbb{R}$. In the domain $\mathbb{R}^D$ the functional combination of $N$ splines are a set of hyperplanes with slope parameters $\alpha \in \mathbb{R}^{DxN}$ and offsets $\beta \in \mathbb{R}^N$. As only one hyperplane is selected for any particular point in the domain this is equivalent to partitioning the domain into N regions. This problem requires the creation of complex polyhedra where each hyperplane meets another~\cite{Chu2018PWLNN, Misener2010PWLapprox, Bunel2020branchPWL}, termed knots in 1-D. If the function is convex then this step can be skipped and the maximum the hyperplanes can taken, avoiding the need for knots. This was exploited by Balestriero et al.~\cite{Balestriero2018spline, Balestriero2021MASO} to create a Max-Affine Spline (MAS) representation of a neural network layer. However, this relies on the use of PWL activation functions and some guarantees need to be made about convexity of the network function. Theoretical explanation is given to how non-convex and non-piecewise function could be approximated with a combination of MASs but no method for generating that was given.

\subsection{A convex-concave representation}
It has been proven than any (PWL) function can be expressed as the difference between two convex PWL linear functions, alternatively the addition of a convex and concave function~\cite{Wang2004cavexPWL}. As any continuous function can be approximated arbitrarily closely by a PWL function as the number of linear elements tends to infinity, by extension this means any continuous function can be approximated by the difference between two convex PWL functions. As was discussed in the MaxOut paper~\cite{Goodfellow2013Maxout} and by Balestriero et al.~\cite{Balestriero2018spline, Balestriero2021MASO} this means the difference between two MASs can approximate any continuous function. When applied to neural network approximation the difficulty then becomes finding the appropriate convex and concave function. 

The Concave-Convex Procedure~\cite{Yuille2003CCCP} was developed to find a suitable combination by minimising an energy function. In this method the Legendre transform is used to move to a suitable domain in which a cost function can be constructed to minimise over. The work of this paper aims to skip the need for optimisation by creating a creating a convex and concave representation by exploiting knowledge of the second derivative of the function that is being approximated.

\section{Methodology}
% Define splines
% Define mas
% Define Legendre transform
% Show any function can be a combination convex and concave
% Explain how to create a convex and concave function
% How it translates to NN
% What to do with ReLU
% Experiments: 
    % toy examples
    % MNIST (training, Hessian, clustering, pytorch)

\subsection{Max-Affine Splines}
Approximation theory aims to approximate a complex function with a simpler one. A spline function is an example of how a complex function can be approximated by the combination of linear segments. A max-affine spline (MAS) is a multivariate function~\cite{DeBoor1978splines} mapping $\mathbb{R}^D \rightarrow \mathbb{R}$ which combines splines using the maximum function, taking the form below
\begin{equation}
    % \mathrm{mas}(x) = \max_{i=1,...,N} \langle \alpha_i \cdot x + \beta_i \rangle,
    \mathrm{mas}(x) = \max_{i=1,...,N} \alpha_i \cdot x + \beta_i,
\end{equation}
where $\alpha_i \in \mathbb{R}^D$ is one of the $N$ slope parameters, $\beta_i \in \mathbb{R}$ is its corresponding offset and $x \in \mathbb{R}^D$ are the input values. As $N \rightarrow \infty$ this can approximate any convex function. Swapping the maximum function to a minimum allows to approximate any concave function.

\subsection{The Legendre transform}
The Legendre transform is a powerful tool used to transform a function of one quantity into a conjugate quantity. This can be used to convert from a function of points $x$ to a function of gradients $m$. The Legendre transform of a function $f(x)$ to a gradient formulation is shown in Eq.~\ref{eq:Legendre}, where $\sup$ is the supremum. 
\begin{equation}\label{eq:Legendre}
    f^*(m) = \sup_{x \in \mathbb{R}} mx - f(x)
\end{equation}
By sampling from $f^*(m)$ a collection of gradients can be used to describe the original function. They take the form of a linear function $y=mx+c$ where $m$ is the gradient and $c=f(x)$. The maximum operation can be used in place of the supremum and then this takes the form a MAS. The problem with taking a Legendre transform of a neural network, as is the aim of this work, is that it only works for convex functions.

\subsection{A convex-concave representation of any function}\label{sec:ccf}
\textbf{Theorem 1:} Any function with a well-defined second derivative can be expressed as the sum of a convex function and a concave function~\cite{Klee1976convexconcaveproof}.

\textbf{Theorem 2}: Any continuous PWL function can be expressed as the combination of a convex and concave PWL function~\cite{Wang2004cavexPWL}.

For a function to be convex the second derivative must be greater than or equal to zero, $f''(x) \geq 0$. This means if the second derivative of a function is know a convex representation can be constructed by adding a term to the second derivative to ensure it is always above zero. Using the sigmoid function as an example 
\begin{equation}
    f(x) = \frac{1}{1 + e^{x}} = \sigma(x),
\end{equation}
the second derivative is
\begin{equation}
    f''(x) = \sigma(x)(1 - \sigma(x))(1 - 2 \sigma(x)),
\end{equation}
which is minimum at $x = \frac{1}{2}+\frac{1}{2 \sqrt{3}}$ with $f''(x) = \frac{-1}{6 \sqrt{3}} \approx -0.96$. By adding $0.96$ to the second derivative and twice integrating back a convex function is created. Any constant from integration can be set to zero as it does not effect the second derivative.
\begin{equation}
    f_{\text{convex}}(x) = \frac{1}{1 + e^{x}} + 0.96x^2,
\end{equation}
however, the original function is lost because of the additional $x^2$ term. We can follow a similar method to create a concave representation of the function which can cancel out the $x^2$ term,
\begin{equation}
    f_{\text{concave}}(x) = \frac{1}{1 + e^{x}} - 0.96x^2, 
\end{equation}
which leads to
\begin{equation}
    f(x) = 0.5 (f_{\text{convex}} + f_{\text{concave}}).
\end{equation}
Due to the symmetry of the sigmoid function the minimum of the second derivative is the opposite sign of the maximum of the second derivative. In other more complicated functions the maximum magnitude of the maximum and minimum of the second derivative must be added to create complementary convex and concave parts. This leads to the general form
\begin{equation}
     f(x) = 0.5 (f(x) + cx^2 +f(x) - cx^2),
\end{equation}
where
\begin{equation}
    c = \max (|\max(f''(x))|, |\min(f''(x))|).
\end{equation}
By splitting the function into a convex and concave part the Legendre transform can now be applied to create two complementary MASs approximations of the original function. In the implementation throughout this work $f(x)$ and $cx^2$ are kept separate to avoid any washing out of values of $f(x)$ in floating point due to large values of $cx^2$. It requires no more parameters than keeping a separate convex and concave function as $c$ is common across both.

\subsection{A convex-concave neural network}
A neural network with appropriate neuron activation is a bounded function with a well-defined second derivative. As $\mathbf{x}$ is now a vector the first step requires taking the Hessian of the overall network. This creates a Hessian matrix of partial derivatives $H = \nabla^2f(\mathbf{x})$. The original function is convex if and only if $H$ is positive semi-definite (PSD) for all $\mathbf{x}$. The Hessian is symmetric since
\begin{equation}
    \frac{\delta}{\delta x_i\delta x_j}f(\mathbf{x}) = \frac{\delta}{\delta x_j\delta x_i}f(\mathbf{x}),
\end{equation}
and a symmetric matrix is PSD if and only if the eigenvalues are positive.

\textbf{Proposition 1:} If $\nabla^2f(\mathbf{x})$ entries are bounded, there exists $c > 0$ such that $H + cI$ is PSD for all $\mathbf{x}$.

If all eigenvalues are positive it is already convex and $c=0$. Assuming the minimum eigenvalue, $\varepsilon$, is less than zero $c$ is non-zero. To calculate $c$ let $\lambda$ be an eigenvalue of $H$ with eigenvector $\mathbf{v}$ then
\begin{equation}
    \lambda \mathbf{v} = H \mathbf{v},
\end{equation}
\begin{equation}
    \lambda \mathbf{v} + c \mathbf{v} = H \mathbf{v} + c \mathbf{v},
\end{equation}
\begin{equation}
    (\lambda + c) \mathbf{v} = (H + c) \mathbf{v},
\end{equation}
so $\lambda + c$ is an eigenvalue of $H + c$. Then set $c = -\varepsilon$ and the Hessian is now PSD. But what is the convex function of $\nabla^2f(\mathbf{x}) + cI$? Consider 
\begin{equation}
    g(\mathbf{x}) = \mathbf{x}^T\mathbf{x} = \sum_i \mathbf{x}^2,
\end{equation}
\begin{equation}
    \frac{\delta}{\delta x_i\delta x_j}g(\mathbf{x}) = \begin{cases}
        \text{1} & \text{if  $i = j$} \\
        0 & \text{otherwise,} 
    \end{cases}
\end{equation}
so the Hessian of $g(\textbf{x}) = I$. Therefore $\nabla^2f(\mathbf{x}) + cI$ is the Hessian of $f(\mathbf{x}) + c\mathbf{x}^T\mathbf{x}$. Similar to the method described in Section~\ref{sec:ccf} the minimum value of the eigenvalue is used for the convex and maximum for the concave function. The maximum absolute value is used to generate a common value of $c$ which guarantees that both partitions of the neural network are complementary and respectively convex and concave. This procedure is carried out for each output of the network individually, creating a value of $c$ for each.
\begin{figure*}[btp]
    \centering
    \includegraphics[width=0.8\textwidth]{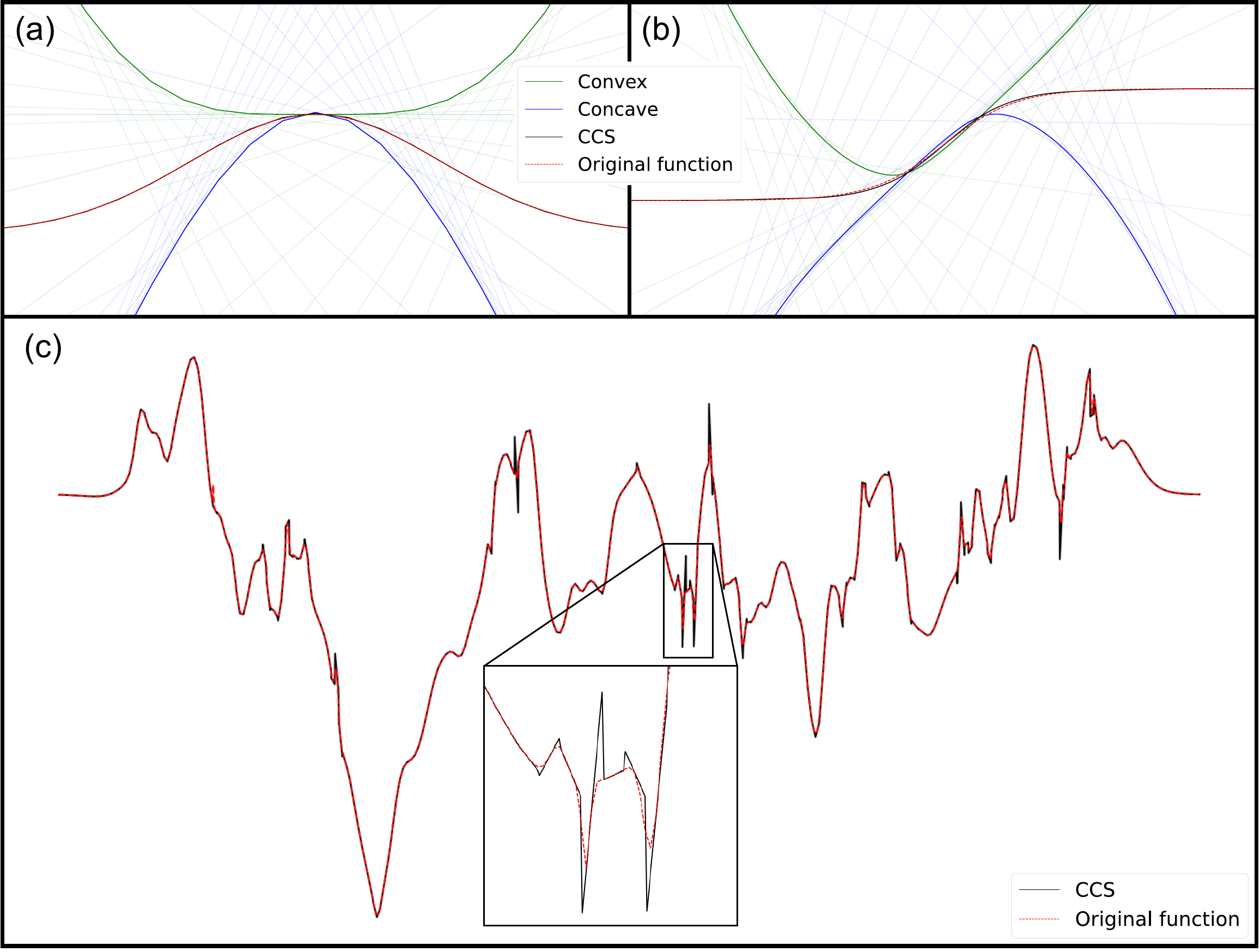}
    \caption{(a) A reconstruction of a Gaussian function with CCS. The faint lines show the planes of which the maximum (minimum) is taken to produce the convex (concave) function.
    (b) The same as (a) for a sigmoid activation function.
    (c) The CCS of 400 Gaussian functions combined. The inset shows the largest approximation error coming from limited sampling around a period of rapid change.}
    \label{fig:Legendre_1D}
\end{figure*}

Practically, when performing this step, as calculating the exact form of the Hessian at all points is computationally complex, the Hessian is evaluated at each input data point using the PyTorch Hessian function. The minimum and maximum value across all points is used to generate the value of $c$ for each output.

\subsubsection{A convex-concave PWL neural network}
% As functions like ReLU do not have a well-defined second derivative a convex alternate form cannot be generated. It is twice differentiable but because the value at $x=0$ is not defined its second derivative is not well-defined, which is a requirement of the transformation. 
Functions, such as ReLU, are twice differentiable but because of their piecewise nature their second derivative is not well-defined which is a requirement when generating the Hessian.
Experimentally it was found that using the value of $c$ calculated by a neural network using non PWL functions trained on the same task was sufficient. It will also be shown later that relatively arbitrary values of $c$ can be used to create approximate convex and concave forms of PWL neural networks. 

% \subsection{Approximating non-convex functions}

\subsection{Approximating MNIST}
The MNIST dataset with the standard 60,000/10,000 train/test split is used for benchmarking neural network approximation~\cite{LeCun1998mnist}. 0.5 is subtracted from input values to centre them around zero. All networks in the following section were trained in PyTorch with stochastic gradient descent with momentum. Across all examples a learning rate of 0.05 used with a momentum value 0.9 and 50\% dropout trains the network for 300 epochs with cross-entropy loss. Sigmoid activation is used unless otherwise stated. The CNN used uses two convolutional layers followed by a hidden layer of size 200. The first had 1 input channel, 16 output channels, a kernel size of 5x5, stride 1, and padding 2. Sigmoid activation and max pooling (kernel size 2x2) followed. The second convolution layer had 16 input channels, 32 output channels, and the same kernel size, stride, padding, activation, and max pooling.

% Clustering is performed on the extracted planes to explore network complexity. This justification 
The justification for using clustering to explore network complexity
comes from each spline capturing elements of the original function. Therefore, 
the fewer planes required the simpler the function.
% there is a relation because cluster
% By using clustering the planes it is possible to see 
When clustering the gradients and offsets of splines are concatenated into a single vector and clustered using the fast PyTorch K-means package. The clustering is run 10 times at each value of K to attain a mean and standard deviation. All experiments are run using a single NVIDIA a100 80GB GPU. Code for the experiments can be found at \href{https://github.com/adamgoodtime/Legendre_Net.git}{https://github.com/adamgoodtime/Legendre\_Net.git}.

\section{Experiments and Results}
% \begin{itemize}
%     \item Toy example converting a Gaussian/sinewave
%     \item More complex classification to extract features (breast cancer or some other UCI)
%     \item Comparison of different network's planes on MNIST
%     \item Effect of k clusters on performance for different network types
%     \item Feature differences across clusters
% \end{itemize}

% Approximation of Gaussian and sigmoid
% The reason for limitations with approximation of combination
% mnist approx
% clustering planes
% difference i layers and size and cnn
% extracted features
\subsection{Transforms in 1D}
The first example displays the capability to create a Convex-Concave-Spline (CCS) representation of non-convex functions. Figure~\ref{fig:Legendre_1D}(a) shows the convex and concave pair which are constructed to approximate a Gaussian function. The faintly coloured lines show the uniformly sampled planes with the maximum of the green lines being used to create the convex component and the minimum of the blue lines used to create the concave component. These are then averaged to create the CCS approximation of the Gaussian function. Figure~\ref{fig:Legendre_1D}(b) shows the same procedure to approximate a sigmoid function.

\begin{table*}[]
    \centering
    \begin{tabular}{|c|c|c|c|}
         \hline
         Architecture & Test acc. \% & CCS acc. \% & Approx. loss \%\\ [0.5ex] 
         \hline\hline
        200x5 & 98.47 & 97.69 & -0.78 \\ 
         \hline
        200x4 & 98.53 & 97.94 & -0.59 \\ 
         \hline
        200x3 & 97.9 & 97.35 & -0.55 \\ 
         \hline
        200x2 & 98.11 & 97.44 & -0.67 \\ 
         \hline
        1600x1 & 98.27 & 97.62 & -0.65 \\ 
         \hline
        800x1 & 98.18 & 97.64 & -0.54 \\ 
         \hline
        400x1 & 98.24 & 97.52 & -0.72 \\ 
         \hline
        200x1 & 98.1 & 97.74 & -0.36 \\ 
         \hline
        CNN & 99.04 & 97.81 & -1.23 \\ 
         \hline
        ReLU-200x2 & 98.5 & 97.64 & -0.86 \\ 
         \hline
        ReLU-200x1 & 98.12 & 97.18 & -0.94 \\ 
         \hline
        ReLU-200x1 c=5 & 98.12 & 97.11 & -1.01 \\ 
         \hline
    \end{tabular}
    \caption{The MNIST test accuracy of the final models and the test accuracy of the CCS approximation using all planes. The final column shows the loss in accuracy following approximation. The architecture denotes the hidden layer width and depth using sigmoid activation unless otherwise stated. c=5 is a middling value of $c$ for the ReLU network, although there is little variance around the chosen value of $c$.}
    % \label{tab:acc}
\end{table*} \label{tab:acc}

Figure~\ref{fig:Legendre_1D}(c) applies the same technique to a combination of 400 Gaussian functions with mean, standard deviation and weight drawn uniformly from $[-3, 3]$, $[0, 0.2]$ and $[-1, 1]$ respectively. 300 planes are uniformly sampled across the 1D domain, for both the convex and concave function, which corresponds to the same number of parameters as is used by the Gaussian combination. The error visible at various points is a result of the sampling frequency. The planes are sampled uniformly across the input space and, therefore, if the original function changes rapidly between sample locations the change in gradient is not precisely captured. This can be remedied by increasing the number of planes but this would require more computation resource. There is also the opportunity to non-uniformly sample planes and increase sampling around periods with large second derivatives, however, this is non-trivial.

\subsection{Approximating a network trained on MNIST}
Table~\ref{tab:acc} shows the final test accuracy and approximation accuracy across a range of architectures. The approximation accuracy is relatively consistent across architecture settings. The most significant inaccuracy is from the single layer ReLU networks. Although, little difference is observed when using a uniform value of $c$ or when using the calculated $c$ from the 200x1 sigmoid network. The main difference in approximation loss is a result of the test accuracy of the network rather than the approximation accuracy as that is relatively consistent.

\subsection{Exploring network complexity with clustering}
The CCS approximation generates a plane for every data point in the training set. 
This leads to a perfect correspondence in training accuracy between the original network and the approximate form but effects the testing accuracy. It can also lead to many planes, 60,000 for MNIST. To examine the redundancy of the CCS procedure planes are clustered using K-means. A range of K values is chosen to see how varying the number of planes changes the faithfulness of the approximation. 

\begin{figure*}
    \centering
    \includegraphics[width=0.8\textwidth]{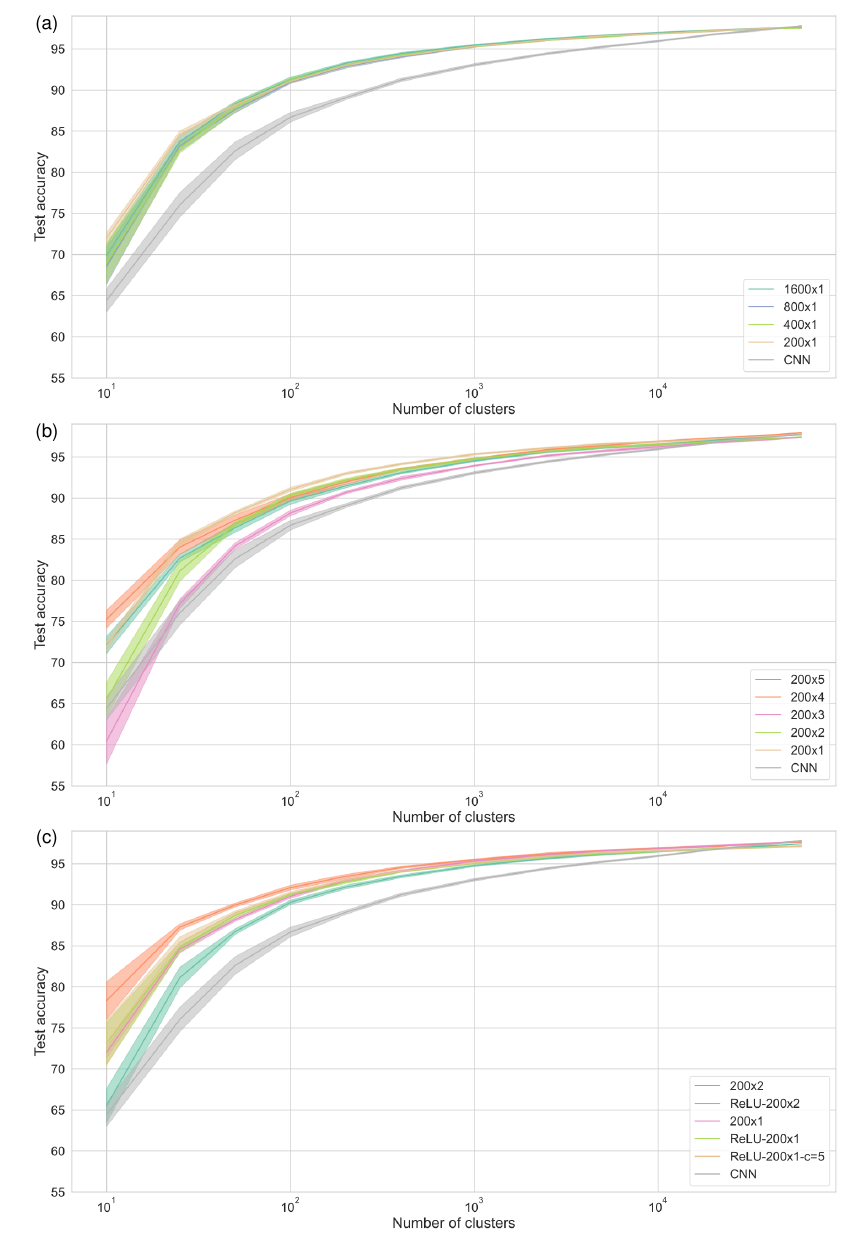}
    \caption{K-means was run until convergence across all examples and then the classification accuracy on the test set was calculated using the selected planes. Clustering was performed 10 times with the line showing the mean and the shaded region the standard deviation. The effect of clustering the CNN planes is shown across all plots as a benchmark. (a) shows the effect of layer width in a neural network with one hidden layer. (b) keeps the layer width the same but alters the depth of the network. (c) displayed how using approximate forms of $c$ for ReLU networks effects approximation and contrasts it with sigmoid networks of the same network size.}
    \label{fig:clustering}
\end{figure*}

Across the plots of Figure~\ref{fig:clustering} the CNN approximation is consistently lower, requiring around a magnitude more clusters to reach similar levels of approximation accuracy compared to other architectures. This can be understood as a measure of the complexity of the network. More planes are required match to function of the network meaning there is less redundancy. This also means the the network varies more across the input space and therefore is more computationally complex. This idea will be further explored in Section~\ref{sec:feat}.

It can be seen in Figure~\ref{fig:clustering}(a) that layer size has little effect on the ability to approximate the network. This is likely a result of the expressivity of the network being fairly consistent across various layer sizes with added neurons not significantly effecting the test accuracy. This means the CCS is capturing a similarly shaped function and therefore achieves similar performance across different hidden layer sizes.

When comparing the approximation accuracy across different sizes of depth in Figure~\ref{fig:clustering}(b) a less clear picture emerges with the shallowest and deepest networks being the best approximated and layer depths of two and three containing the highest approximation errors. A possible explanation for this is that a shallow network has the lowest complexity and therefore can be approximated relatively well. As the number of layers increases the the complexity grows and with it the the algorithm struggles to approximate the complex function. However, once the depth goes beyond a certain point a degree of smoothing happens over the input space as a result of more information dispersal across the layers and regularisation via dropout. This will be explored in the Sec~\ref{sec:feat} where features will be extracted from various architectures and contrasted.

Examining the ReLU network approximation error shows that the choice of $c$ does not need to be precise with a uniform and calculated value of $c$ achieving similar approximation error. The value of $c=5$ was chosen as the uniform value, although there was little sensitivity to the value between the range of 2 and 10. Less than 2 and the convex and concave components were not curved enough to capture the function of the network and beyond 10 there was diminishing returns from increased curvature due to floating point limitations. A surprising result from Figure~\ref{fig:clustering}(c) is that the best approximation came from ReLU network of depth 2 and the worst from a sigmoid network of depth 2. A possible explanation for this is the increased classification of the 2-layer ReLU network. Also, there is no significant increase in classification accuracy between the 2-layer sigmoid network and the 1-layer networks but there will be an increased complexity over the input space which will be harder for the approximation to capture.

\subsection{Examining features}\label{sec:feat}
\begin{figure*}[btp]
    \centering
    \includegraphics[width=1\textwidth]{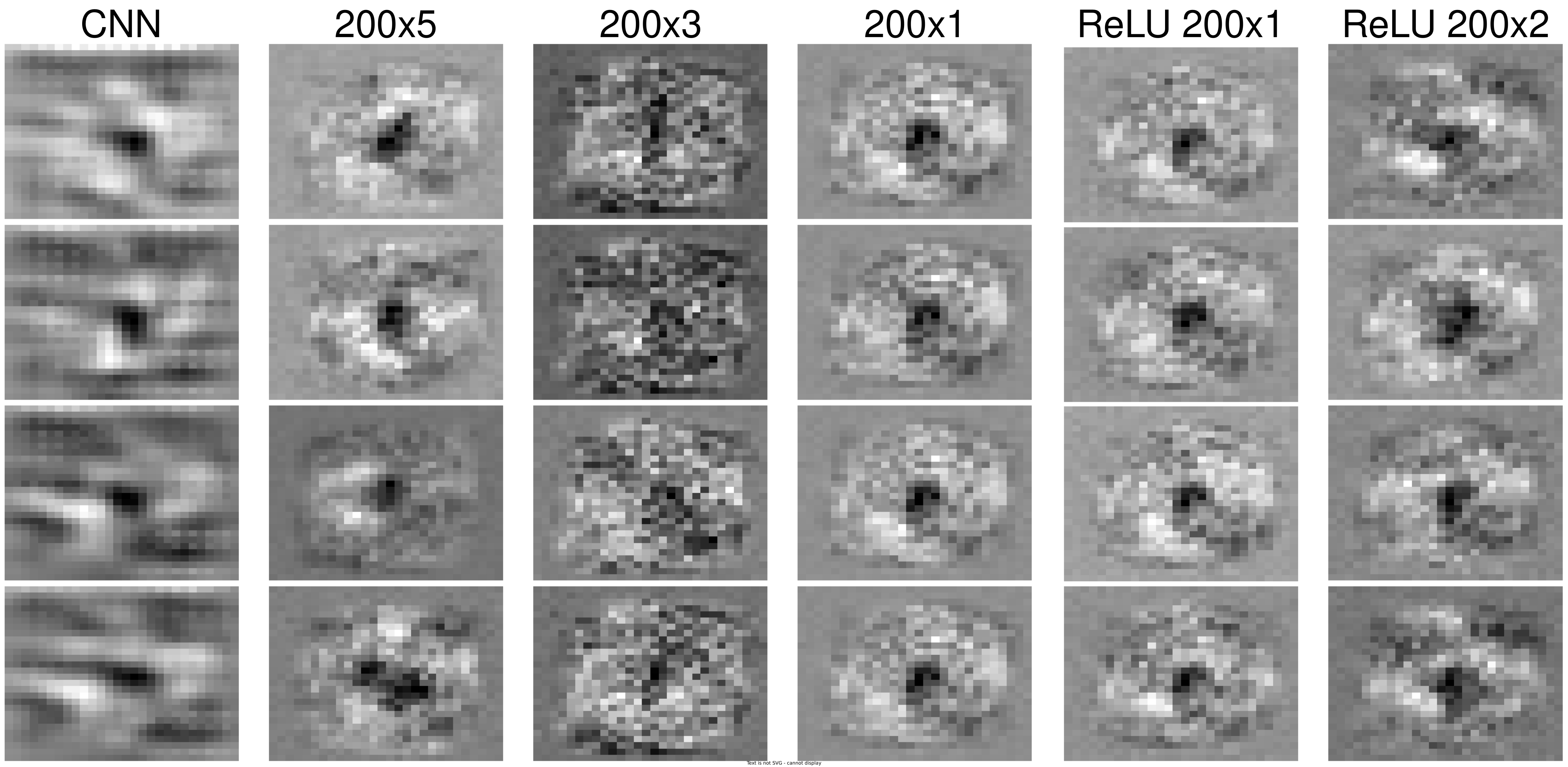}
    \caption{The feature maps of different architectures for the MNIST class zero extracted from the CCS representation. A lighter shade corresponds to a positive weight and darker negative.}
    \label{fig:features}
\end{figure*}
Following the clustering of planes random clusters are selected to inspect the features of the network. A small value of $K=10$ is used to increase the disparity between clusters. As was discussed in Section~\ref{sec:ccf}, the gradient and offset of $f(x)$ and $cx^2$ are kept separate, allowing the extraction of the gradient component of the plane that relates to the original function, $f'(x)$. This corresponds to the network's feature map for a particular output as the gradient shows what input would create the highest variability in response of the network.

Figure~\ref{fig:features} compares the extracted features of four randomly selected clusters for the output zero. The features of shallow networks are consistent, with little variation across clusters. The deeper networks contain noticeable variation across clusters suggesting a higher information content captured by the network function. The 3-layer network has a significantly noisier feature map, which corresponds to the worse test accuracy. Although the 5-layer network has varied feature maps, the approximation is still good due to the quality of the features, which can be seen by their relative smoothness and lack of noise. The CNN has by far the smoothest and most varied feature maps. The features also look the least like the number zero. This suggests a high degree of variability in the CNN response across the range of possible inputs. This explains the need for more planes to approximate the CNN function as there are fewer redundant planes and a diverse set of responses. The smoothness of the response also displays a low sensitivity to noise.

\section{Discussion and Limitations}
% \begin{itemize}
%     \item Way of extracting features which trigger outputs without GD
%     \item Rough metric of network complexity
%     \item Converts net to alternate form which is highly parallelisable/understandable
% \end{itemize}
% H is only at x so not guaranteed for all
% exploding values, FP wash out
% sample limitation
% inspecting individual neurons
It has been shown in this work that an alternate convex and concave representation of any bounded function with a well-defined second derivative can used to construct an approximate form using Max-Affine Splines (MAS), creating a Convex-Concave Spline (CCS) representation. This was displayed for complex non-convex functions and also applied to neural network approximation. This alternate form creates further bridges to approximation theory. It was also shown experimentally that approximate forms can be created for piecewise linear (PWL) functions which possess undefined second derivatives at their points of inflection. 

The CCS representation allowed inspection of feature maps. This enabled the inspection and comparison of network function and an approximation of functional complexity. The construction of the MAS also lends itself to highly parallel computation as each spline can be computed independently. Networks with multiple layers and convolutional filers can be flattened into a combination of linear operators. In theory, the same method can be used for individual neurons, although this is left for future work.

%%%%%%%%%%%%%%%%%%%%%%%%%%%%%%%%%%%%%%%%%%%%%%%%%%%%%%%%%%%
%% shown that CNN (and multi layer NN) features vary across the input space which brings into question probing mechanics

% further optimisation of the splines

% quantified method for network complexibility interpretation

% \begin{enumerate}
%     \item A novel method of splitting a neural network into complementary convex and concave components
%     \item The application and subsequent sampling of a Legendre transform of the convex and concave components to create a max-affine spline approximation of the original neural network that is embarrassingly parallel in operation
%     \item A novel method for feature visualisation and interpretation is proposed
%     \item The investigation of the CCS approximate form to evaluate network complexity
% \end{enumerate}

Although CCS can model any continuous bounded function with a well-defined second derivative it is still a PWL approximation and the approximation error only tends to zero as the number of planes tends to infinity. In this method a sample is taken at each training data point, which lead to an exact correspondence between the network and the CCS in training data accuracy but lead to test accuracy dropping by around 0.5-1\%. In the future, optimisation could be performed to create a more condensed CCS form. Also the Hessian is only evaluated around the training data values which means it may not capture the full potential dynamic range of the network. An exact form was created for the the 1D examples, however, this is left to future work for neural network approximation.

% Given a large enough domain and high variance in the target function there is also the potential for the additional $cx^2$ term to explode, reaching the limits of floating point representation. If a more conservative convex and concave form can be created it may be able to limit the potential of this problem. 
The addition to the second derivative to guarantee convexity is not an optimal choice as $x^2$ can become very large in comparison to the original function. When integrating back the constant was set to zero, choosing a different value could alleviate this problem.
In the work of Yuille and Rangarajan~\cite{Yuille2003CCCP} an energy function is minimised to generate complementary convex and concave functions, although this has not been extended to neural network approximation. 

Comparison across architectures enabled the interrogation of network complexity. As is often the problem with features map generation the interpretation is qualitative and left to subjective explanation. Future work would benefit from quantitative descriptions of the extracted feature maps to make precise and objective claims about the networks being investigated. A measure of variability across feature maps would be straightforward to implement. Designing a metric to measure noise present in a feature map would be more complicated when extending beyond image classification.

% \section{Supplementary Material}

% Authors may wish to optionally include extra information (complete proofs, additional experiments and plots) in the appendix. All such materials should be part of the supplemental material (submitted separately) and should NOT be included in the main submission.

% \section*{References}

% References follow the acknowledgments in the camera-ready paper. Use unnumbered first-level heading for
% the references. Any choice of citation style is acceptable as long as you are
% consistent. It is permissible to reduce the font size to \verb+small+ (9 point)
% when listing the references.
% Note that the Reference section does not count towards the page limit.
\medskip

{
\small

% \printbibliography
\bibliography{myrefs}
\bibliographystyle{icml2023}

%%%%%%%%%%%%%%%%%%%%%%%%%%%%%%%%%%%%%%%%%%%%%%%%%%%%%%%%%%%%

\end{document}